\documentclass{esannV2}
\usepackage[dvips]{graphicx}
\usepackage[latin1]{inputenc}
\usepackage{amssymb,amsmath,array}

\usepackage{booktabs}

\usepackage{hyperref}

\usepackage{subcaption}
\usepackage{caption}
\usepackage{comment}
\usepackage{todonotes}

\begin{document}

\title{Noise Robust One-Class Intrusion Detection on Dynamic Graphs}

\author{Aleksei Liuliakov, Alexander Schulz, Luca Hermes, and Barbara Hammer
%
\thanks{We gratefully acknowledge funding by the BMBF within the project HAIP, grant number 16KIS1212 and by SAIL, which is funded by the Ministry of Culture and Science of the State of North Rhine-Westphalia under the grant no NW21-059A.}
%
\vspace{.3cm}\\
%
Bielefeld University - Machine Learning Group \\
CITEC building, Inspiration 1, 33619 Bielefeld - Germany
%
}

\maketitle

\begin{abstract}
In the domain of network intrusion detection, robustness against contaminated and noisy data inputs remains a critical challenge. This study introduces a probabilistic version of the Temporal Graph Network Support Vector Data Description (TGN-SVDD) model, designed to enhance detection accuracy in the presence of input noise. By predicting parameters of a Gaussian distribution for each network event, our model is able to naturally address noisy adversarials and improve robustness compared to a baseline model. Our experiments on a modified CIC-IDS2017 data set with synthetic noise demonstrate significant improvements in detection performance compared to the baseline TGN-SVDD model, especially as noise levels increase. Our implementation is available online.\footnote{\url{https://github.com/AlekseiLiu/rtgn_svdd}}

\end{abstract}

\section{Introduction}
Network intrusion detection systems (NIDS) play a crucial role in protecting information infrastructures from various cyber threats. Machine learning techniques are becoming increasingly popular and sophisticated in the field of NIDS, particularly unsupervised methods that enable the detection of novel attacks \cite{kocher2021machine, singh2021machine, mustafa2024intrusion}. Numerous factors can affect the performance of machine learning-based NIDS. One important factor is noise within the input signal \cite{al2021empirical}, which can originate from a variety of sources. For instance, attackers might intentionally introduce adversarial noise to obfuscate real attacks and mislead the system. Data poisoning or contamination is a prevalent strategy used to deceive ML models by manipulating the data \cite{nkashama2022robustness}. Furthermore, limitations in hardware and software related to capturing, processing, and storing real-world network data often lead to multiple irregular faults and errors, thereby increasing signal noise. 
These challenges highlight the need for robust detection mechanisms capable of overcoming such adversities. Traditional models, however, often struggle with the noisy and dynamic nature of network data, degrading their performance.  

Recent advances in graph neural networks (GNN) offer promising avenues for capturing complex relationships in network data through dynamic graph structures \cite{liuliakov2023one}. However, these models generally rely on clean and precise input data and lack resilience to noise. Addressing this, we propose a probabilistic extension to the TGN-SVDD model \cite{liuliakov2023one}. This novel approach does not only detect intrusions and attacks but also explicitly targets unknowns caused by noisy inputs by predicting the parameters of a Gaussian distribution. This dual capability allows the model to effectively manage the uncertainty in noisy network environments. We evaluate our model using a popular open source data set CIC-IDS2017 \cite{Sharafaldin2018TowardGA}, which has been artificially augmented with various types of noise to simulate real-world conditions. Our results highlight the enhanced robustness and reliability of our approach compared to the baseline method.

\section{Fundamentals}
In this work we focus on internet traffic as a data source. Traffic can be represented as a set of Network flows \cite{hofstede2014flow}. These can be consider as a continuous-Time Dynamic Graph (CTDG), where each network flow corresponds to an edge addition event in the graph with source IP and destination IP as source and destination nodes, respectively. Network flow's statistics features become an edge features in CTGN. Further we follow the same notation and key concept definitions as were used in the work \cite{liuliakov2023one}.

\subsection{Continuous-Time Dynamic Graphs (CTDG)}\label{sec:ctgd}

Temporal (multi-)graphs are a sequences $G = \{x(t_1), x(t_2), ...\}$ of time-stamped events $x(t)$, which we consider to be interaction events in the form of directed temporal edges $\mathbf{e}_{ij}(t)$, which may be accompanied by an edge feature vectors $\mathbf{f}_{ij}(t)\in\mathbb{R}^{f}$. Let $E(T) = \{(i, j) : \exists\, \mathbf{e}_{ij}(t) \in G, t \in T\}$ be an edge set, we define $F(T) = \{i: \exists (i, .)\in E(T) \lor \exists (., i)\in E(T)\}$ as a set of all nodes up to time $T$.

\subsection{The TGN-SVDD}
In the work \cite{liuliakov2023one} the TGN-SVDD framework was introduced, which is an unsupervised approach to anomaly detection in Dynamic Graphs. It combines the temporal encoding capabilities of the TGN with the anomaly detection framework of Deep SVDD.

The \textbf{TGN Encoder}
generates temporal node embeddings by maintaining a memory vector for each node, which is updated using a learnable memory function (e.g., GRU) after each interaction event. Each node $i$ has a memory state $\mathbf{h}_i(t_{k}^-)$, updated whenever node $i$ participates in an event and $t_{k}^-$ is the timestamp prior to $t_{k}$. The TGN encoder function $\mathbf{z}(i, \mathbf{h}_i(t_{k}^-), \mathcal{N}_i(t_k), \mathbf{W})$ generates an embedding for node $i$ based on its prior state $\mathbf{h}_i(t_{k}^-)$, its temporal neighborhood $\mathcal{N}_i(t_k)$, and encoder parameters $\mathbf{W}$, mapping it into an embedding space $\mathcal{F} \subseteq \mathbb{R}^{p}$. For the sake of conciseness, we denote the TGN encoder as $\mathbf{z}_i(t_k, \mathbf{W})$.

The \textbf{Deep SVDD Decoder} is adapted to work with the TGN encoder's output. It calculates an anomaly score by measuring the squared Euclidean distance between the concatenated embeddings of interacting nodes (from TGN) and a trainable center $\mathbf{c}$ of a hypersphere in the embedding space: 
$s(x(t_k), \mathbf{W}, \mathbf{c}) = \lVert\, \left(\mathbf{z}_i(t_k, \mathbf{W})\oplus\mathbf{z}_j(t_k, \mathbf{W})\right) - \mathbf{c} \,\rVert^2$, with $i$ and $j$ being event node's indexes at time $t_k$ and $\oplus$ the vector concatenation operation.

The \textbf{Optimization Objective} for the TGN-SVDD model aims to minimize the anomaly score, alongside a regularization term for the encoder parameters: $\min_{\mathbf{W},\mathbf{c}} ~ \frac{1}{D} \sum_{k = 1}^D  s(x(t_k), \mathbf{W}, \mathbf{c}) + \lambda \lVert \,\mathbf{W} \,\rVert^2$, with $D$ being the total number of training interaction events which belongs to the single (normal) class of the data set and $\lambda$ the regularization hyperparameter controlling the weight decay.

\section{Proposed model: Robust TGN-SVDD}

\textbf{Problem Formulation:}
We consider the dynamic setting described in Section \ref{sec:ctgd}, where we now identify three types of events: normal events, malicious attacks, and events attributable to noise. We assume access only to normal events during training, with the goal being to classify attacks against the combined class of normal and noise events. We assume that noise consists of irregular stochastic CTDG events, $x(t)$. Dynamic edge events induced by pairs $(i, j) | i, j \sim \text{Uniform}(F(T))$, where $T$ is the maximum timestamp of the dataset, and the edge features $\mathbf{f}_{ij}(t) \sim \mathcal{N}(0, \Sigma)$, characterized by the covariance matrix $\Sigma$, are randomly added.

We will see that such noise affects the performance of deterministic TGN-SVDD. We propose a novel unsupervised IDS for Dynamic Graphs via a probabilistic extension of TGN-SVDD, that is robust against random and irregular events at inference.

\noindent\textbf{Proposed Modeling:} We extend the TGN-SVDD from a point-wise regression task to a probabilistic regression task. We employ a multidimensional Gaussian distribution as a predicted distribution for a given input. Thus, given an input event $x(t_k)$ with a corresponding node pair $(i, j) \in E(t_k)$, the TGN encoder outputs two vectors, $\mathbf{\mu_i}(t_k, \mathbf{W})$ and $\mathbf{\sigma_i}(t_k, \mathbf{W})$, for each node, representing the mean and variance of the Gaussian distribution, respectively. Our model assumes the following relationship for the output parameters: $\mathbf{\mu_i}(t_k, \mathbf{W}) \oplus \mathbf{\mu_j}(t_k, \mathbf{W}) \sim \mathcal{N}(\boldsymbol{c}, \mathbf{diag}(\mathbf{\sigma_i}(t_k, \mathbf{W}) \oplus \mathbf{\sigma_j}(t_k, \mathbf{W})))$. This suggests the choice of an objective as Negative Log Likelihood (NLL), which takes form:

\begin{align*}
\min_{\mathbf{W},\mathbf{c}} \quad & \frac{1}{D\cdot N} \sum_{k=1}^D \sum_{m=1}^N \bigg( \log\left(\left(\sigma_{im}(t_k, \mathbf{W})\oplus\sigma_{jm}(t_k, \mathbf{W})\right)^2\right) \\
& +  \frac{(z_{im}(t_k, \mathbf{W})\oplus z_{jm}(t_k, \mathbf{W}) - c_m)^2}{(\sigma_{im}(t_k, \mathbf{W})\oplus\sigma_{jm}(t_k, \mathbf{W}))^2} \bigg),
\end{align*}
\noindent where N is the dimension of the events embedding. The covariance is a diagonal matrix, NLL decomposes into a sum of NLL for each dimension.
We propose two scores and a two-fold decision process:
\begin{itemize}
\item Apply the score $s_{\sigma}(\mathbf{x}(t_k)) = \frac{1}{N} \sum_{m=1}^N \left(\sigma_{im}(t_k, \mathbf{W}) \oplus \sigma_{jm}(t_k, \mathbf{W})\right)$ to distinguish noisy exemplars from the rest. The larger $s_{\sigma}(\mathbf{x}(t_k))$ is, the more likely $\mathbf{x}(t_k)$ is to be noise.
\item Use $s_{\mu}(\mathbf{x}(t_k)) = \lVert\, \left(\mathbf{z}_i(t_k, \mathbf{W})\oplus\mathbf{z}_j(t_k, \mathbf{W})\right) - \mathbf{c} \,\rVert^2$ as anomaly score to detect attacks in the rest of the data.
\end{itemize}
\noindent\textbf{Negative sampling:}
In unsupervised learning settings where the model is exposed only to the normal class, we incorporate negative sampling during the training phase to enhance the responsiveness of $s_{\sigma}(\mathbf{x}(t_k))$ to noise. For each training iteration, we obtain negative examples by sampling node indexes $(v, w) | v, w \sim \text{Uniform}(F(T))$ for the input events. Different sampling strategies can be applied, which may vary depending on assumptions about the noise source. In this work, we apply random uniform sampling with replacement from the complete set of possible nodes $F(T)$, where $T$ represents the maximum time in the training dataset.
For these negative examples, we modify the objective:

\begin{align*}
\min_{\mathbf{W}} \quad & \frac{1}{D\cdot N} \sum_{k=1}^D \sum_{m=1}^N \bigg( \log\left(\left(\hat{\sigma}_{vm}(t_k, \mathbf{W}) \oplus \hat{\sigma}_{wm}(t_k, \mathbf{W})\right)^2\right) \\
& + \frac{(\hat{\mu}_{vw})^2}{\left(\hat{\sigma}_{vm}(t_k, \mathbf{W}) \oplus \hat{\sigma}_{wm}(t_k, \mathbf{W})\right)^2} \bigg),
\end{align*}
\noindent where we use $\hat{\sigma}_{vm}$ for negative samples. We sample the negative mean value as $\hat{\mu}_{vw} \sim \mathcal{N}(0, \Sigma)$, where $\Sigma$ is a hyperparameter that amplifies the noise effect for negative examples. Both positive and negative objectives are simultaneously optimized during training.

\section{Experiment}
In this section, we introduce the data, specify the experimental setup, and report comparative metrics from the experiments.

\noindent\textbf{Data set:}
We employ the CIC-IDS2017 dataset \cite{Sharafaldin2018TowardGA} for model evaluation. This dataset includes realistic network intrusion scenarios and consists of both normal and malicious traffic collected over a workweek. We preprocess the data into four separate datasets, each split into training, validation, and test subsets. The initial segments contain only normal traffic, with attacks reserved for testing. More details about the data and its preprocessing steps can be found in \cite{liuliakov2023one}.

\noindent\textbf{Noise:}
We craft and inject noise events into the test sets as follows: each communication event contains corresponding source and destination node IDs $(i, j)$, a feature vector $\mathbf{f}_{ij}$, and a timestamp $t$. $i, j \sim \text{Uniform}(F(T))$, $\mathbf{f}_{ij} \sim \mathcal{N}(0, \text{diag}(5))$, and $t \sim \text{Uniform}([t_{test}, T])$, where $t_{test}$ is the timestamp of the first test event, and $T$ is the maximum timestamp in the dataset. We simply add such random events to the test set for further experiments and evaluations.

\noindent\textbf{Experimental Setup:}
To tune and adjust hyperparameters, we held out one dataset, specifically the Wednesday working hours with 5\% noise injected into the test set. The binary classification task was to discriminate between attacks and the combined class of normal and noise events. The TGN-SVDD was trained over 30 epochs. For each epoch, the F1 score has been evaluated for a variety of thresholds. The choice, which corresponds to the best-performing epoch was used for further experiments. A similar procedure was used to tune the proposed RTGN-SVDD model. A set of F1 scores was computed based on a grid of threshold pairs. During this procedure, the following negative sampling parameters were determined: the amount of negative samples was set to 30\%  of the positive examples, and $\Sigma$ was used for the covariance Gaussian distribution. Although the model performs well for various $\Sigma$ values, we chose $\Sigma = \text{diag}(5)$ as it performs best within our setup.  We also used the dataset to determine the interval of the best-performing thresholds for anomaly score $s_{\mu}$, which in our experimental setting is $[5, 25]$.

\noindent\textbf{Results:}
The evaluation task for both TGN-SVDD and RTGN-SVDD is to classify the attack class against the combined class of normal and attack. The single baseline evaluation uses a ROC-AUC metric calculated from TGN's $s$ output. For the proposed model, we iterate over each possible threshold for $s_{\sigma}$ within the interval $[5, 25]$ and compute the ROC-AUC for $s_{\mu}$. The mean value of this set of ROC-AUC scores contributes to a single evaluation. We repeat the noise resampling for each level of noise and each day five times to obtain mean and standard deviation values over single evaluations of both models. These final summary values are reported in the table \ref{Tab:roc_auc_summary}.

\begin{figure}[h]
\centering

\includegraphics[width=0.49\textwidth]{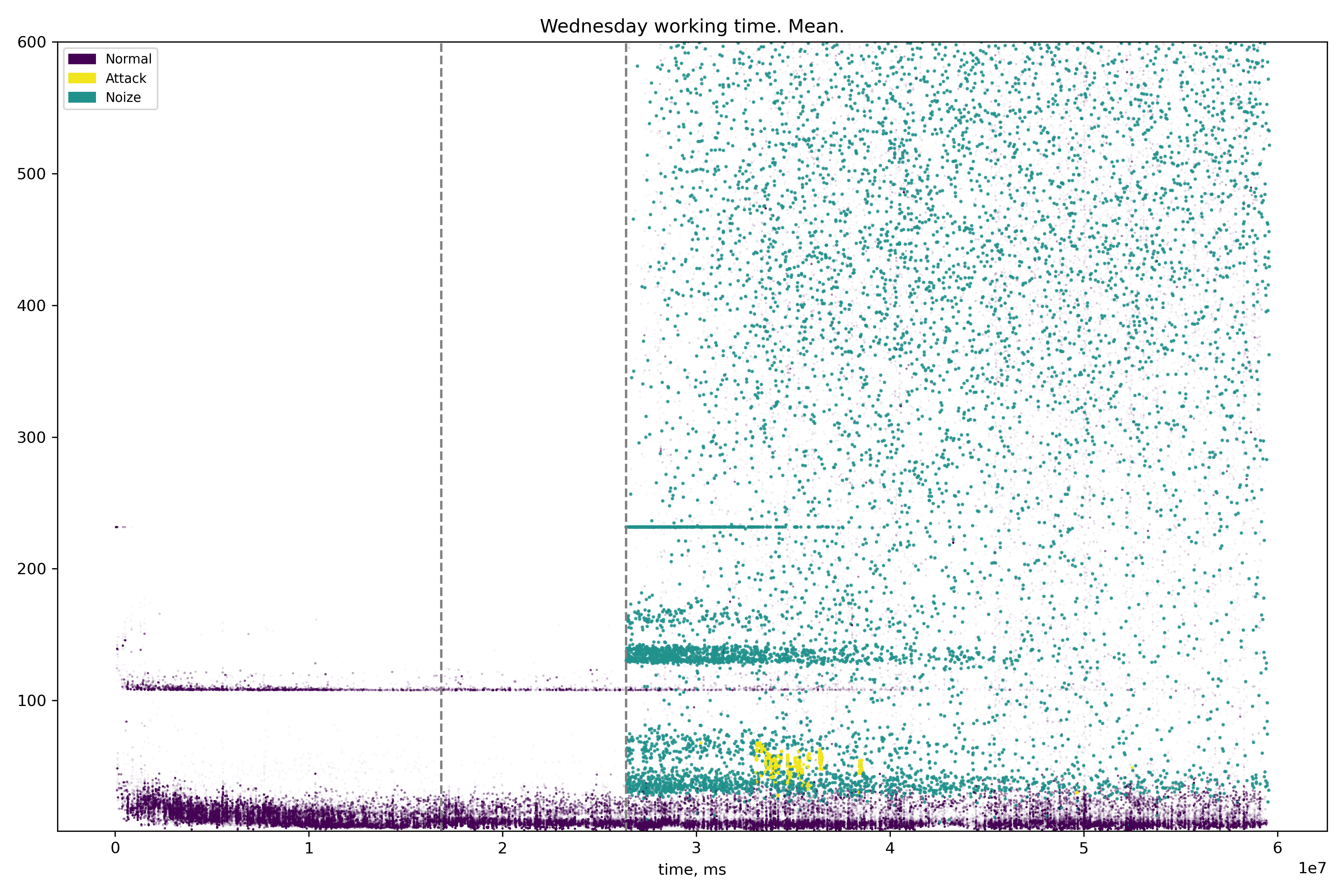}
\includegraphics[width=0.49\textwidth]{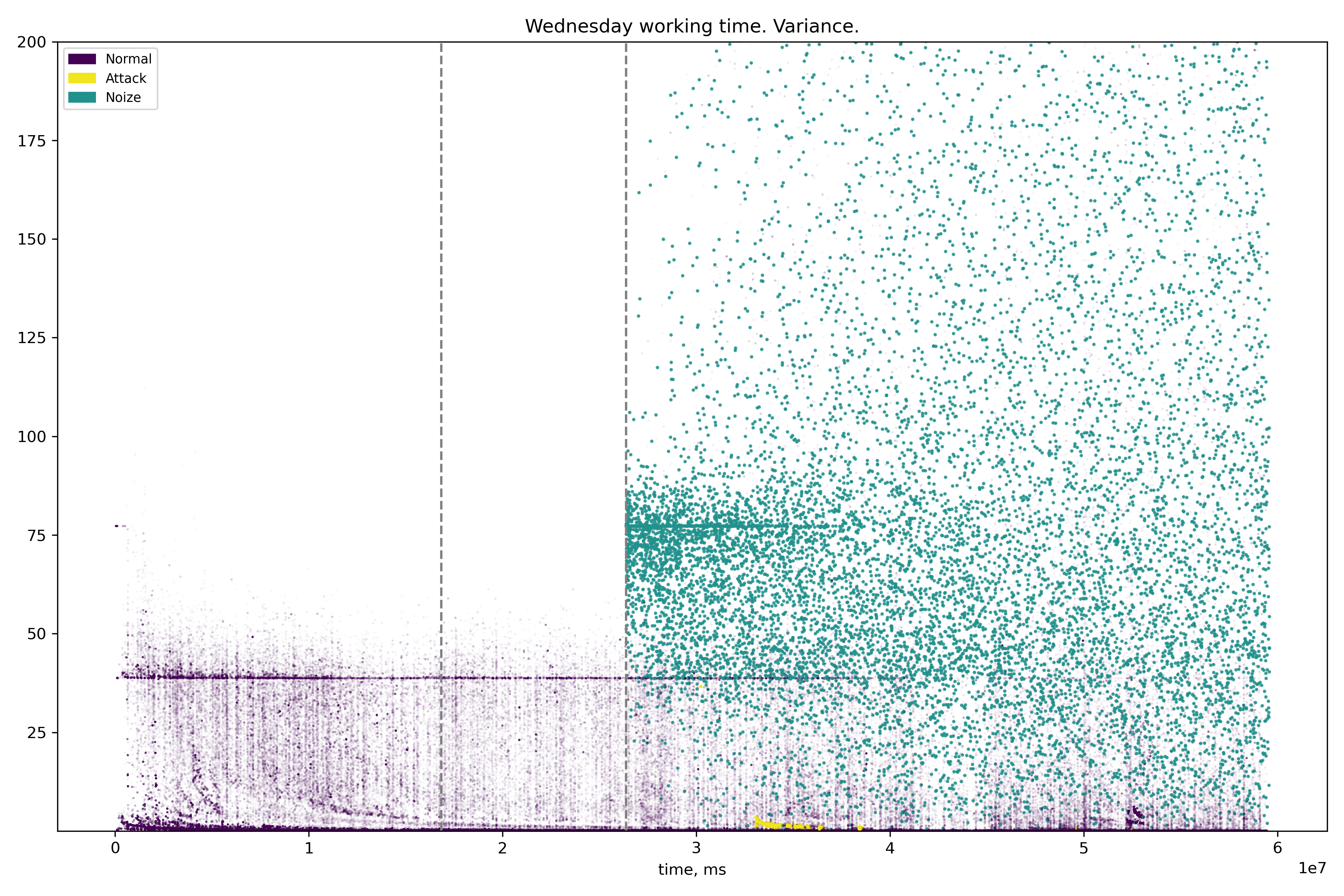}
\caption{Illustration of RTGN-SVDD performance using the Wednesday working hours dataset, with a noise level of 5\% in the test set. Dashed vertical lines separate the training, validation, and test sets respectively. On the right: the $s_{\mu}$ score output; on the left: the $s_{\sigma}$ score output.}\label{Fig:rtgn-mean-sigma}
\end{figure}

To illustrate the performance of RTGN-SVDD, we provide plots \ref{Fig:rtgn-mean-sigma} of the scores per event versus time: $s_{\mu}$ on the left and $s_{\sigma}$ on the right side.

Results suggest that our proposed model significantly outperforms the baseline model in the case of inference in a noisy environment, showing ever-increasing robustness with the rise in noise levels compared to the baseline.
\begin{table}[h!]
  \centering
  \normalsize 
  \setlength{\tabcolsep}{3pt} 
  \caption{Summary of ROC-AUC Metrics for TGN-SVDD and RTGN-SVDD on Tuesday, Thursday and Friday working hours data sets respectively. The table corresponds to attack against combined normal and noise class classification task. First column contains the ratio in \% of the amount of the noise to the amount of the normal examples. All the rest columns contain corresponding ROC-AUC summary metrics multiplied by 100.}
  \begin{tabular}{ccccccc}
  \toprule
     & \multicolumn{2}{c}{Tuesday} & \multicolumn{2}{c}{Thursday} & \multicolumn{2}{c}{Friday} \\
     
    \midrule
    Noise & TGN  & RTGN & TGN  & RTGN & TGN  & RTGN    \\
    \midrule
10 & $81.5 \pm 2.0$ & $\mathbf{87.1} \pm 2.3$& $80.3 \pm 9.7$ & $77.1 \pm 5.4$ & $83.0 \pm 0.5$ & $\mathbf{92.7} \pm 0.1$\\
20 & $72.6 \pm 1.5$ & $\mathbf{85.5} \pm 2.7$ & $66.7 \pm 7.0$ & $\mathbf{76.6} \pm 5.6$  & $73.3 \pm 0.7$ & $\mathbf{91.1} \pm 0.3$ \\
30 & $65.8 \pm 1.5$ & $\mathbf{85.1} \pm 2.1$ & $69.5 \pm 4.7$ & $\mathbf{79.5} \pm 3.9$ & $67.0 \pm 0.9$ & $\mathbf{90.0} \pm 0.4$ \\
40 & $59.9 \pm 2.6$ & $\mathbf{84.7} \pm 1.7$ & $63.2 \pm 2.9$ & $\mathbf{79.4} \pm 6.5$ & $60.9 \pm 1.2$ & $\mathbf{89.1} \pm 0.3$ \\
50 & $57.4 \pm 0.6$ & $\mathbf{84.4} \pm 1.5$ & $61.2 \pm 1.7$ & $\mathbf{73.7} \pm 6.3$ & $56.1 \pm 1.3$ & $\mathbf{88.0} \pm 0.3$ \\
    \bottomrule
  \end{tabular}
  \label{Tab:roc_auc_summary}
\end{table}

\section{Conclusion}
We introduced an enhanced TGN-SVDD model that integrates probabilistic methods to improve resilience against noisy data in NIDS. We demonstrated significant detection accuracy improvements under noisy conditions using the modified CIC-IDS2017 dataset, highlighting the model's potential in adversarial network environments. Future work could explore the application of this model to additional datasets and its application of the noise quantitative output for integration with alternative modality detection systems to further validate and enhance its effectiveness.



\begin{footnotesize}



\bibliographystyle{unsrt}
\bibliography{bibliography_test.bib}

\begin{thebibliography}{1}

\bibitem{kocher2021machine}
Geeta Kocher and Gulshan Kumar.
\newblock Machine learning and deep learning methods for intrusion detection
  systems: recent developments and challenges.
\newblock {\em SC}, 25(15):9731--9763, 2021.

\bibitem{singh2021machine}
Richa Singh, Nidhi Srivastava, and Ashwani Kumar.
\newblock Machine learning techniques for anomaly detection in network traffic.
\newblock In {\em 2021 6 (ICIIP)}, volume~6, pages 261--266. IEEE, 2021.

\bibitem{mustafa2024intrusion}
Zaid Mustafa, Rashid Amin, Hamza Aldabbas, and Naeem Ahmed.
\newblock Intrusion detection systems for software-defined networks: a
  comprehensive study on machine learning-based techniques.
\newblock {\em Cluster Computing}, pages 1--27, 2024.

\bibitem{al2021empirical}
Khalid~M Al-Gethami, Mousa~T Al-Akhras, and Mohammed Alawairdhi.
\newblock Empirical evaluation of noise influence on supervised machine
  learning algorithms using intrusion detection datasets.
\newblock {\em Security and Communication Networks}, 2021:1--28, 2021.

\bibitem{nkashama2022robustness}
D~Nkashama, Arian Soltani, Jean-Charles Verdier, Marc Frappier, Pierre-Marting
  Tardif, and Froduald Kabanza.
\newblock Robustness evaluation of deep unsupervised learning algorithms for
  intrusion detection systems.
\newblock {\em arXiv preprint arXiv:2207.03576}, 2022.

\bibitem{liuliakov2023one}
Aleksei Liuliakov, Alexander Schulz, Luca Hermes, and Barbara Hammer.
\newblock One-class intrusion detection with dynamic graphs.
\newblock In {\em ICANN}, pages 537--549. Springer, 2023.

\bibitem{Sharafaldin2018TowardGA}
Iman Sharafaldin, Arash~Habibi Lashkari, and A.~Ghorbani.
\newblock Toward generating a new intrusion detection dataset and intrusion
  traffic characterization.
\newblock In {\em ICISSP}, 2018.

\bibitem{hofstede2014flow}
Rick Hofstede, Pavel {\v{C}}eleda, Brian Trammell, Idilio Drago, Ramin Sadre,
  Anna Sperotto, and Aiko Pras.
\newblock Flow monitoring explained: From packet capture to data analysis with
  netflow and ipfix.
\newblock {\em IEEE Communications Surveys \& Tutorials}, 16(4):2037--2064,
  2014.

\end{thebibliography}

\end{footnotesize}

\end{document}